\documentclass[10pt,twocolumn,letterpaper]{article}

\usepackage{cvpr}              

\usepackage{capt-of}
\usepackage[dvipsnames]{xcolor}
\usepackage{xspace}
\usepackage{enumitem}
\usepackage{amsmath,amssymb,amsbsy,amsfonts,dsfont,pifont,bm,bbm,mathrsfs,mathtools,nicefrac}
\usepackage{algorithm,algpseudocode,listings}
\usepackage{booktabs,multirow,adjustbox,diagbox,threeparttable,tabularray}
\usepackage[accsupp]{axessibility}

\newcommand{\method}{AnyDoor\xspace}

\newcommand{\tocite}[1]{{\color{red} [TO CITE]}}

\newcommand{\ve}[1]{\mathbf{#1}} 

\definecolor{cvprblue}{rgb}{0.21,0.49,0.74}
\usepackage[pagebackref,breaklinks,colorlinks,citecolor=cvprblue]{hyperref}
\usepackage{wrapfig}
\usepackage[capitalize]{cleveref}  
\crefname{section}{Sec.}{Secs.}
\Crefname{section}{Section}{Sections}
\crefname{table}{Tab.}{Tabs.}
\Crefname{table}{Table}{Tables}
\crefname{figure}{Fig.}{Figs.}
\Crefname{figure}{Figure}{Figures}
\crefname{equation}{Eq.}{Eqs.}
\Crefname{equation}{Equation}{Equations}
\newcolumntype{x}[1]{>{\centering\arraybackslash}p{#1}}
\newcolumntype{y}[1]{>{\raggedright\arraybackslash}p{#1}}
\newcolumntype{z}[1]{>{\raggedleft\arraybackslash}p{#1}}

\def\paperID{1362}
\def\confName{CVPR}
\def\confYear{2024}

\title{\method: Zero-shot Object-level Image Customization}

\author{
    Xi Chen$^{1}$ \quad
    Lianghua Huang$^{2}$ \quad
    Yu Liu$^{2}$ \quad
    Yujun Shen$^{3}$ \quad
    Deli Zhao$^{2}$ \quad
    Hengshuang Zhao$^{1*}$\\[5pt]
    $^{1}$The University of Hong Kong \quad
    $^{2}$Alibaba Group \quad
    $^{3}$Ant Group
}

\begin{document}
\twocolumn[{
\renewcommand\twocolumn[1][]{#1}
\maketitle
\begin{center}
    \vspace{-20pt}
    \includegraphics[width=1.0\linewidth]{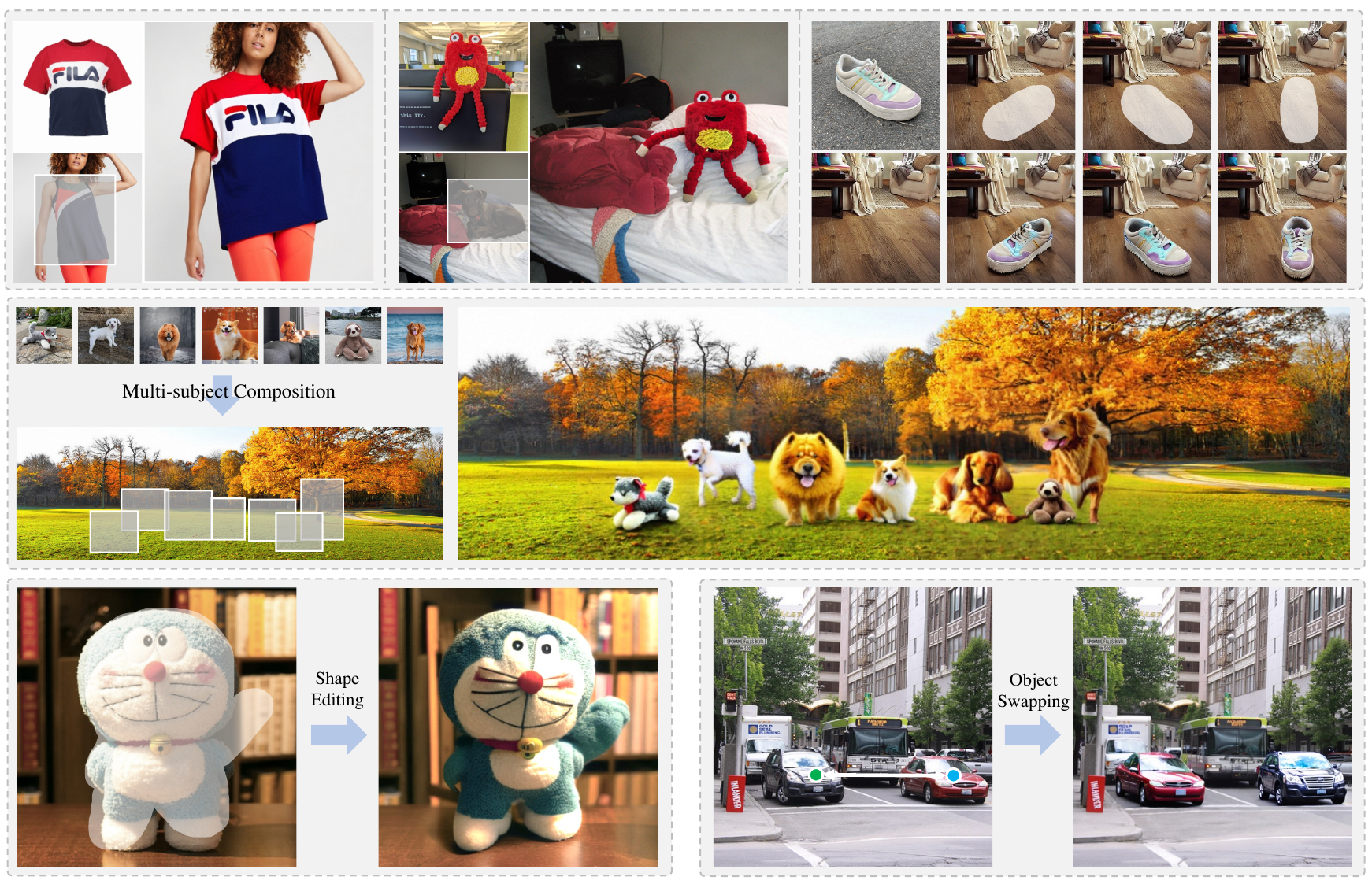}
    \vspace{-20pt}
    \captionsetup{type=figure}
    \caption{%
        \textbf{Fantastic applications} of our proposed \method \textit{without} any parameter tuning.
        Our model allows users to customize an image by placing an object at specific locations, with optional shape controls (top row).
        It can be extended to handle multiple objects (middle row) and also supports seamlessly editing the shape of the object or swapping objects within real scenes (bottom row).
    }
    \label{fig:teaser}
    \vspace{10pt}
\end{center}
}]

\let\thefootnote\relax\footnotetext{*Corresponding author.}
\begin{abstract}

This work presents \textbf{\method}, a diffusion-based image generator with the power to teleport target objects to new scenes at user-specified locations
with desired shapes.
Instead of tuning parameters for each object, our model is trained only once and effortlessly generalizes to diverse object-scene combinations at the inference stage.
Such a challenging zero-shot setting requires an adequate characterization of a certain object.
To this end, we complement the commonly used identity feature with detail features, which are carefully designed to maintain appearance details yet allow versatile local variations (\textit{e.g.}, lighting, orientation, posture, \textit{etc.}), supporting the object in favorably blending with different surroundings.
We further propose to borrow knowledge from video datasets, where we can observe various forms (\textit{i.e.}, along the time axis) of a single object, leading to stronger model generalizability and robustness.
Extensive experiments demonstrate the superiority of our approach over existing alternatives as well as its great potential in real-world applications, such as virtual try-on, shape editing, and object swapping.
Code is released at \href{https://github.com/ali-vilab/AnyDoor}{github.com/ali-vilab/AnyDoor}.
\end{abstract}
\vspace{-12pt}

\section{Introduction}
Image generation is flourishing with the booming advancement of diffusion models~\cite{ldm,imagen,controlnet,composer,DALLE2,unicontrol}. Humans could generate favored images by giving text prompts, scribbles, skeleton maps, or other conditions. The power of these models also brings the potential for image editing. For example, some works~\cite{imagic,masactrl,sine} learn to edit the posture, styles, or content of an image via instructions. 
Other works~\cite{inpaintanything,smartbrush} explore re-generating a local image region with the guidance of text prompts.

In this paper, we investigate ``object teleportation'',  
which means accurately and seamlessly placing the target object into the desired location of the scene image.
Specifically, we re-generate a box/mask-marked local region of a scene image by taking the target object as the template.  This ability is a significant requirement in practical applications, like image composition, effect-image rendering, poster-making, virtual try-on, \textit{etc}.

Although strongly in need, this topic is not well explored by previous researchers. Paint-by-Example~\cite{paintbyexample} and Objectstitch~\cite{objectstitch} take a target image as the template to edit a specific region of the scene image, but they could not generate ID~(identity)-consistent contents, especially for untrained categories. Customized synthesis methods~\cite{dreambooth,textinversion,liu2023cones,cones2,multiconcept} are able to conduct generations for the new concepts but could not be specified for a location of a given scene. Besides, most customization methods need finetuning on multiple target images for nearly an hour, which largely limits their practicability for real applications.

We address this challenge by proposing \method. Different from previous methods, \method is able to generate ID-consistent compositions with high quality in zero-shot.  To achieve this,  we represent the target object with identity- and detail-related features, then composite them with the interaction of the background scene. Specifically, we use an ID extractor to produce discriminative ID tokens and delicately design a frequency-aware detail extractor to get detail maps as a supplement. We inject the ID tokens and the detail maps into a pre-trained text-to-image diffusion model as guidance to generate the desired composition.
To make the generated content more customizable, we explore leveraging additional controls~(\textit{e.g.} user-drawn masks) to indicate the shape/poses of the object.
To learn customized object generation with high diversities, we collect image pairs for the same object from videos to learn the appearance variations, and also leverage large-scale statistic images to guarantee the scenario diversity.

Equipped with these techniques, \method demonstrates extraordinary abilities for zero-shot customization.  As in \cref{fig:teaser}, \method shows promising performance for the synthesis of the new concept with shape controls (top row).
Besides, since \method owns the high controllability for editing the specific local regions of the scene image, it is easy to be extended to multi-subject composition (middle row), which is a hot and challenging topic explored by many customized generation methods~\cite{multiconcept,Mix-of-Show,Break-A-Scene,cones2}. Moreover, the high generation fidelity and quality of \method unlock the possibilities for more fantastic applications like object moving and swapping (bottom row). We hope that \method could serve as a foundation solution for various image generation and editing tasks with image input, and act as the basic ability to energize more fancy applications.
\vspace{-5pt}
\section{Related Work}
\vspace{-5pt}
\noindent\textbf{Local image editing.}
Most of the previous works focus on editing local image regions with text guidance. Blended Diffusion~\cite{blendeddiffusion} conducts multi-step blending in the masked region to generate more harmonized outputs. Inpaint Anything~\cite{inpaintanything} involves SAM~\cite{segmentanything} and Stable Diffusion~\cite{ldm} to replace any object in the source image with text described target. Paint-by-Example~\cite{paintbyexample} uses CLIP~\cite{CLIP} image encoder to convert the target image as an embedding for guidance, thus painting a semantic consistency object on the scene image. ObjectStitch~\cite{objectstitch} proposes a similar solution with ~\cite{paintbyexample}, which trains a content adaptor to align the outputs of the CLIP image encoder to the text encoder to guide the diffusion process. However, those methods could only give coarse guidance for generation and often fail to synthesize ID-consistent results for untrained new concepts.

\definecolor{flamecolor}{RGB}{235,51,35}
\definecolor{snowflakecolor}{RGB}{47,110,186}
\begin{figure*}[t]
\centering 
\includegraphics[width=1.0\linewidth]{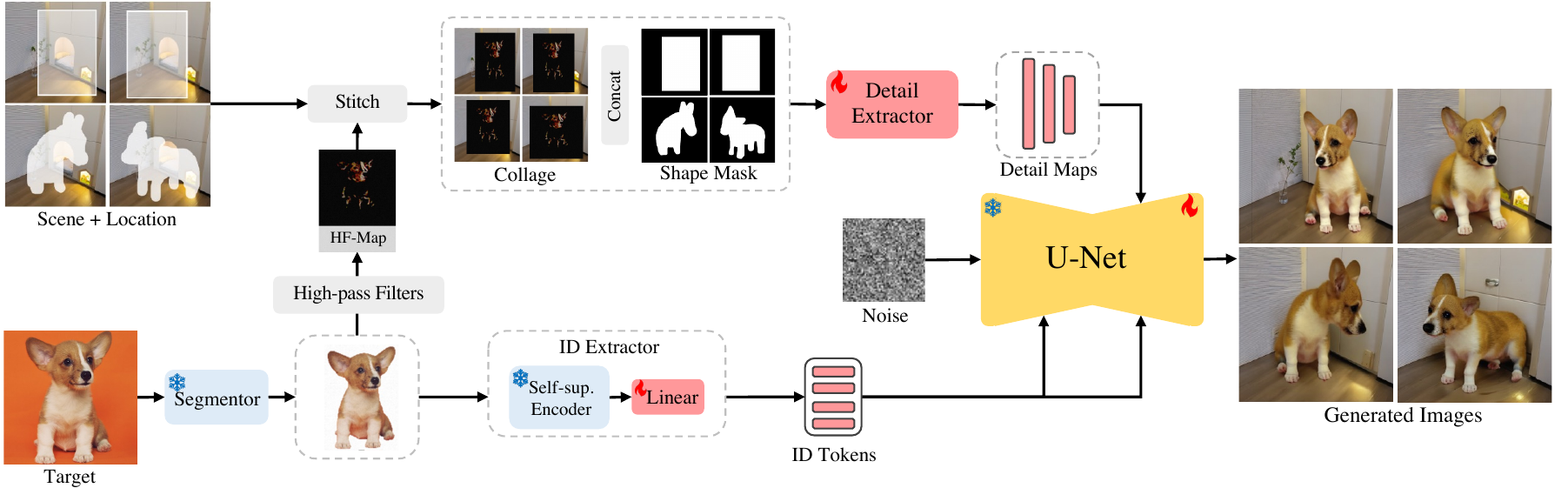} 
\vspace{-15pt}
\caption{%
    \textbf{Overall pipeline} of \method, which is designed to teleport an object to a scene with the desired location and shape.
    A segmentation module first removes the object background, followed by an ID extractor to obtain its identity information (\cref{subsec:id_extractor}).
    We then apply high-pass filters to the ``clean'' object, stitch the resulting high-frequency map (HF-Map) with the scene at the desired location, and concatenate the collage and shape mask.
    A detail extractor is designed to complement the ID extractor with appearance details (\cref{subsec:detail_extractor}).
    Finally, the ID tokens and detail maps are injected into a pre-trained diffusion model to produce the final synthesis, 
    where the target object favorably blends with its surroundings~(\cref{subsec:feature_injection}).
    \textbf{\textcolor{flamecolor}{Flames}} and \textbf{\textcolor{snowflakecolor}{snowflakes}} refer to learnable and frozen parameters, respectively.
}
\label{fig:pipeline}
\vspace{-5pt}
\end{figure*}

\noindent\textbf{Customized image generation.}
Customized or termed subject-driven generation aims to generate images for specific objects given several target images and relevant text prompts. Some works~\cite{textinversion,dreambooth,disenbooth} fine-tune a ``vocabulary'' to describe the target concepts. Cones~\cite{liu2023cones} finds the corresponding neurons for the referred object. Although they could generate high-fidelity images, the user could not specify the scenario and the location of the target object. Besides, the time-consuming finetuning impedes them from being used in large-scale applications. 
Recently, BLIP-Diffusion~\cite{blipdiffusion} leverages BLIP-2~\cite{blip2} to align images and text for zero-shot customization. 
Fastcomposer~\cite{fastcomposer} binds the image representation with certain text embeddings to do multiple-person generation.  
Some concurrent works~\cite{ye2023ipadapter,yuan2023customnet,li2023dreamedit} also explore using one reference image to customize the generation results but fail to keep the fine details.

\noindent\textbf{Image harmonization.}
A classical image composition pipeline is cutting the foreground object and pasting it on the given background. Image harmonization~\cite{multiharm,dovenetharm,towardharm,intrinsicharm} could further adjust the pasted region for more reasonable lighting and color. DCCF~\cite{dccf} designs pyramid filters to better harmonize the foreground. CDTNet~\cite{dualtrasnharm} leverages dual transformers. HDNet~\cite{hierarchicalharm} proposes a hierarchical structure to consider both global and local consistency and reaches the state-of-the-art.  Nevertheless, these methods only explore the low-level changes, editing the structure, view, and pose of the foreground objects, or generating the shadows and reflections are not taken into consideration.
\section{Method}

The pipeline of \method is demonstrated in \cref{fig:pipeline}. Given the target object, the scene, and the location, \method generates the object-scene composition with high fidelity and diversity.  The core idea is representing the object with identity- and detail-related features, and recomposing them in the given scene by injecting those features into a pre-trained diffusion model. To learn the appearance changes, we leverage large-scale data including both videos and images for training.

\subsection{Identity Feature Extraction}\label{subsec:id_extractor}
We leverage the pre-trained visual encoders to extract the identity information of the target object.  Previous works~\cite{paintbyexample,objectstitch} choose CLIP~\cite{CLIP} image encoder to embed the target object. 
However, as CLIP is trained with text-image pairs with coarse descriptions, it could only embed semantic-level information but struggles to give discriminative representations that preserve the object identity.  To overcome this challenge, we make the following updates.

\noindent\textbf{Background removal.}
Before feeding the target image into the ID extractor, we remove the background with a segmentor and align the object to the image center. The segmentor model could be either automatic~\cite{segmentanything,u2net} or interactive~\cite{focalclick,cdnet,simpleclick}.  This operation has proven helpful in extracting more neat and discriminative features.

\noindent\textbf{Self-supervised representation.}
In this work, we find the self-supervised models show a strong ability to preserve more discriminative features. Pretrained on large-scale datasets, self-supervised models are naturally equipped with the instance-retrieval ability and could project the object into an augmentation-invariant feature space. We choose the currently strongest self-supervised model DINOv2~\cite{dinov2} as the backbone of our ID extractor, which encodes image as a global token $\mathbf{T}_{\text{g}}^{1\times1536}$, and patch tokens $\mathbf{T}_{\text{p}}^{256\times1536}$.  We concatenate the two types of tokens to preserve more information. We find that using a single linear layer as a projector could align these tokens to the embedding space of the pre-trained text-to-image UNet. The projected tokens $\mathbf{T}_{\text{ID}}^{257\times1024}$ are noted as our ID tokens. 

\subsection{Detail Feature Extraction}\label{subsec:detail_extractor}

Considering that the ID tokens are represented in low resolution~($16\times16$) , it would be hard for them to maintain the low-level details adequately. Thus, we need extra guidance for the detail generation in complementary.

\noindent\textbf{Collage representation.}
Inspired by \cite{controllablecollage,Collagediffusion}, using collage as controls could provide strong priors, we attempt to stitch the ``background removed object'' to the given location of the scene image. With this collage, we observe a significant improvement in the generation fidelity, but the generated results are too similar to the given target which lacks diversity. Facing this problem, we explore setting an information bottleneck to prevent the collage from giving too many appearance constraints.  
Specifically, we design a high-frequency map to represent the object, which could maintain the fine details yet allow versatile local variants like the gesture, lighting, orientation, \textit{etc}.

\noindent\textbf{High-frequancy map.}
We extract the high-frequency map of the target object with
\begin{equation}
    \mathbf{I}_h = (\mathbf{I}_{\text{gray}} \otimes \mathbf{K}_h + \mathbf{I}_{\text{gray}} \otimes \mathbf{K}_v) \odot \mathbf{I} \odot \mathbf{M}_{\text{erode}},
    \label{eq:high}
\end{equation}
where $\mathbf{K}_h, \mathbf{K}_v$ denote horizontal and vertical Sobel~\cite{sobel} kernels, acting as high-pass filters. $\otimes, \odot$ refer to convolution and Hadamard product.  Given an Image $\mathbf{I}$, we first extract the high-frequency regions using these high-pass filters, then extract the RGB colors using the Hadamard product. We also add an eroded mask $\mathbf{M}_{\text{erode}}$ to filter out the information near the outer contour of the target object.
%

\begin{figure}[t]
\centering 
\includegraphics[width=1.0\linewidth]{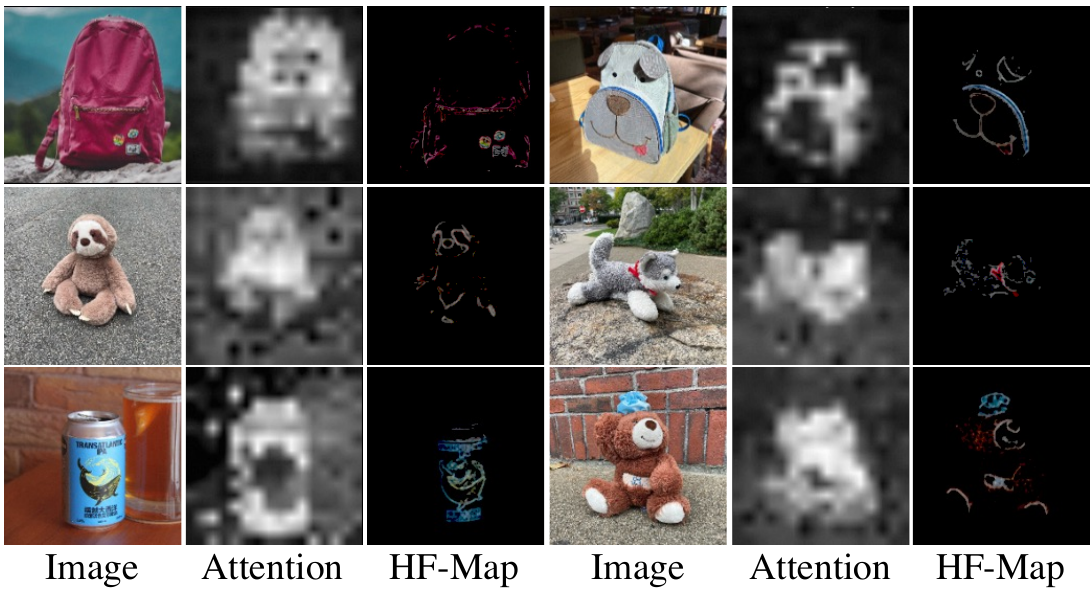} 
\vspace{-18pt}
\caption{%
    \textbf{Visualization of the focus region} of \textit{ID extractor} and \textit{detail extractor}.
    ``Attention'' refers to the attention map of the ID extractor backbone~(DINOv2~\cite{dinov2}), while ``HF-Map'' refers to the high-frequency map used in the detail extractor.
    These two modules focus on global and local information in complementary.
}
\label{fig:highfreq}
\vspace{-10pt}
\end{figure}

%
As visualized in \cref{fig:highfreq}, the tokens produced by DINOv2 focus more on the overall structure, leaving it hard to encode the fine details like the logos of the backpack in the first row. In contrast, the high-frequency map could help take care of these details as a complementary.

\noindent\textbf{Shape control.} We use a shape mask to indicate the object's gestures. To simulate the user input, we downsample the ground truth masks with different ratios and apply random dilation/erosion to remove the details. To keep the ability to tackle a single box input,  we set a probability of 0.3 to use the inner box region as the mask.  During training, the object counter would be aligned with the shape mask. Thus, the users could control the target object's shape by drawing coarse shape masks during inference.

After getting the collage and the contour map, we concatenate them and feed them into the detail extractor.  The detail extractor is a ControlNet-style~\cite{controlnet} UNet encoder, which produces a series of detail maps with hierarchical resolutions.

\begin{figure}[t]
\centering 
\includegraphics[width=1.0\linewidth]{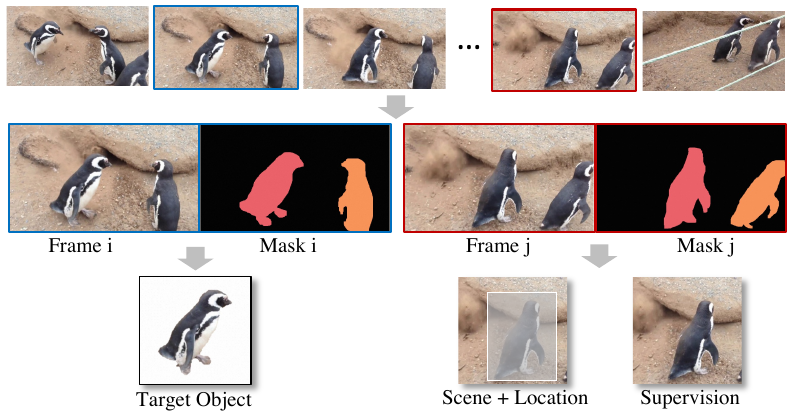} 
\vspace{-15pt}
\caption{%
    \textbf{Data preparation pipeline} for videos.
    Given a clip, we first sample two frames and get the masks for the instances within each frame.
    Then, we select an instance from one frame as the target object and treat the same instance on the other frame as the supervision (\textit{i.e.}, the desired model output).
}
\label{fig:data}
\vspace{-2pt}
\end{figure}

\begin{table}[t]
\caption{%
    \textbf{Statistics of datasets} used for training.
    ``Variation'' refers to whether an object enjoys local variations (\textit{e.g.}, lighting, viewpoint, posture, \textit{etc.}) within a data entry, while ``quality'' particularly refers to image resolution.
}
\label{tab:datasets}
\vspace{-7pt}
\centering\scriptsize
\setlength{\tabcolsep}{5.0pt}
\begin{tabular}{@{}ccccc}
\toprule
\textbf{Dataset}                    & \textbf{Type}    & \textbf{\# Samples} & \textbf{Variation} & \textbf{Quality} \\
\midrule
YouTubeVOS~\cite{youtubevos}        & Video            & 4,453                & \ding{51}          & Low   \\
YouTubeVIS~\cite{youtubevis}        & Video            & 2,883                & \ding{51}          & Low   \\
UVO~\cite{UVO}                      & Video            & 10,337               & \ding{51}          & Low   \\
MOSE~\cite{ding2023mose}            & Video            & 1,507                & \ding{51}          & High  \\
VIPSeg~\cite{vipseg}                & Video            & 3,110                & \ding{51}          & High  \\
BURST~\cite{burst}                  & Video            & 1,493                & \ding{51}          & Low   \\
\midrule
MVImgNet~\cite{mvimgnet}            & Multi-view Image & 104,261             & \ding{51}          & High  \\
VitonHD~\cite{vitonhd}              & Multi-view Image & 11,647              & \ding{51}          & High  \\
FashionTryon~\cite{fahiontryon}     & Multi-view Image & 21,197              & \ding{51}          & High  \\
\midrule
MSRA-10K~\cite{MSRA10K}             & Single Image     & 10,000              & \ding{55}          & High  \\
DUT~\cite{DUT}                      & Single Image     & 15,572              & \ding{55}          & High  \\
HFlickr~\cite{dovenetharm}          & Single Image     & 4,833               & \ding{55}          & High  \\
LVIS~\cite{lvis}                    & Single Image     & 118,287             & \ding{55}          & High  \\
SAM~(subset)~\cite{segmentanything} & Single Image     & 100,864             & \ding{55}          & High  \\
\bottomrule
\end{tabular}
\vspace{-12pt}
\end{table}

\subsection{Feature Injection} \label{subsec:feature_injection}
After getting the ID tokens and detail maps, we inject them into a pre-trained text-to-image diffusion model to guide the generation.  We pick Stable Diffusion~\cite{ldm}, which projects the images into latent space and conducts the probabilistic sampling using a UNet. We note the pre-trained UNet as $\hat{\ve{x}}_{\theta}$, it starts denoising from an initial latent noise $ \ve{\epsilon} \sim \mathcal{U}([0, 1])$ and takes the text embedding $\ve{c}$ as the condition to generate new image latent $ \ve{z}_t = \alpha_t \hat{\ve{x}}_{\theta}(\ve{\epsilon}, \ve{c}) + \sigma_t \ve{\epsilon} $. The training supervision is a mean square error loss as:
\begin{equation}
    \label{euq:ldm}
    \mathbb{E}_{\ve{x}, \ve{c},\ve{\epsilon},t}(\| \hat{\ve{x}}_{\theta}(\alpha_t \ve{x} + \sigma_t \ve{\epsilon}, \ve{c}) - \ve{x} \|^2_2).
\end{equation}
$\ve x$ is the ground-truth image latent, $t$ is the diffusion timestep, $\alpha_t, \sigma_t$ are denoising hyperparameters. 

In this work, we replace the text embedding $\ve{c}$ as our ID tokens, which are injected into each UNet layer via cross-attention. For the detail maps, we concatenate them with UNet decoder features at each resolution. During training, we freeze the pre-trained parameters of the UNet encoder to preserve the priors and tune the UNet decoder to adapt it to our new task.

\begin{figure*}[t]
\centering 
\includegraphics[width=1.0\linewidth]{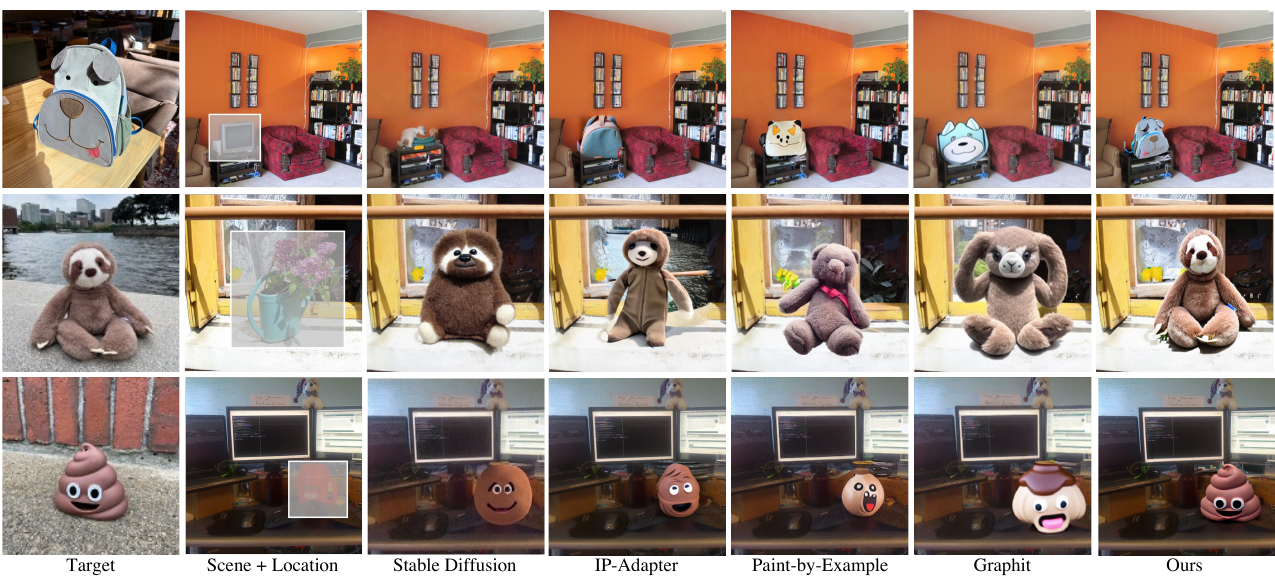} 
\vspace{-17pt}
\caption{%
    \textbf{Qualitative comparison with reference-based image generation methods}, including Stable Diffusion~\cite{ldm}, IP-Adapter~\cite{ye2023ipadapter}, Paint-by-Example~\cite{paintbyexample}, and Graphit~\cite{Graphit}, where our \method better preserves the identity of the target object.
    Note that all approaches do \textit{not} fine-tune the model on the test samples.
}
\label{fig:demobig}
\vspace{-5pt}
\end{figure*}

\subsection{Training Strategies} 
\label{sec:data}

\noindent\textbf{Image pair collection.}
The ideal training samples are image pairs for ``the same object in different scenes'', which are not directly provided by existing datasets. As alternatives,  previous works~\cite{paintbyexample,objectstitch} leverage single images and apply augmentations like rotation, flip, and elastic transforms. However, these naive augmentations could not well represent the realistic variants of the poses and views. 

To deal with this problem, in this work, we utilize video datasets to capture different frames containing the same object. 
The data preparation pipeline is demonstrated in \cref{fig:data}, where we leverage video segmentation/tracking data as examples.  For a video, we pick two frames and take the masks for the foreground object. Then, we remove the background for one image and crop it around the mask as the target object. This mask could be used as the mask control after perturbation.  For the other frame, we generate the box and remove the box region to get the scene image, and the unmasked image could serve as the training ground truth. 
The full data used is listed in \cref{tab:datasets}, which covers a large variety of domains like nature scenes, virtual try-on, saliency, and multi-view objects.

\noindent\textbf{Adaptive timestep sampling.}
Although the video data would be beneficial for learning the appearance variation, the frame qualities are usually unsatisfactory due to the low resolution or motion blur.
In contrast, images could provide high-quality details and versatile scenarios but lack appearance changes.  
To take advantage of both video data and image data, we develop adaptive timestep sampling to make different modalities of data to benefit different stages of denoising training. The original diffusion model~\cite{ldm} evenly samples the timestep~(T) for each training data. However, it is observed that the initial denoising steps mainly focus on generating the overall structure, the pose, and the view, and the later steps cover the fine details like the texture and colors. Thus, for the video data, we increase the possibility by 50\% of sampling early denoising steps~(500-1000) during training to better learn the appearance changes. For images, we increase 50\% probabilities of the late steps~(0-500) to learn how to cover the fine details.
\section{Experiments} \label{sec:exp}

\begin{figure*}[t]
\centering 
\includegraphics[width=1.0\linewidth]{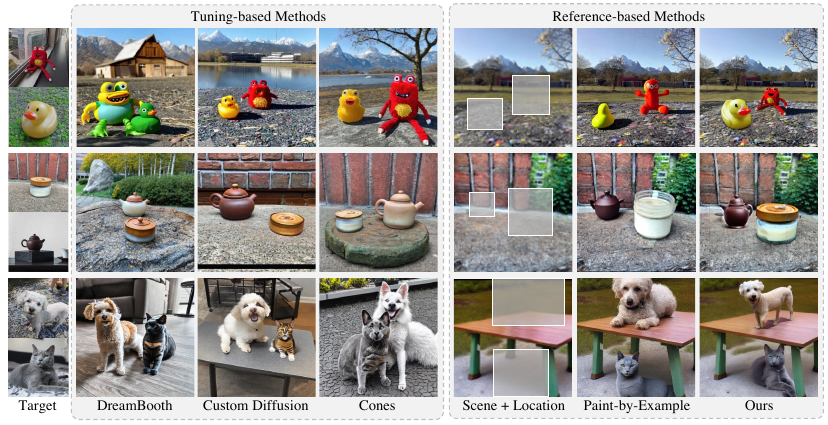} 
\vspace{-17pt}
\caption{%
    \textbf{Qualitative comparisons with existing alternatives for multi-subject composition}, including DreamBooth~\cite{dreambooth}, Custom Diffusion~\cite{multiconcept}, Cones~\cite{liu2023cones}, and Paint-by-Example~\cite{paintbyexample}, where our \method better preserves the object identity and harmoniously blends into the surroundings \textit{without} any parameter tuning.
}
\label{fig:multi}
\vspace{-5pt}
\end{figure*}

\vspace{-6pt}
\subsection{Implementation Details}
\vspace{-6pt}
\noindent\textbf{Hyperparameters.}
We choose Stable Diffusion V2.1~\cite{ldm} as the base generator.  
During training, we process the image resolution to $512\times512$. We choose  Adam~\cite{adam} optimizer with an initial learning rate of $1e^{-5}$. 
We train two versions of models, the original version only takes the box to indicate the location, and the plus version tasks shape masks. In this paper, if not specified with a shape mask, the results are produced by the original version.

\noindent\textbf{Zoom-in strategy.}
During inference, given a scene image and a location box, we expand the box into a square with an amplifier ratio of 2.0. Then, we crop the square and resize it to $512\times512$ as the input for our diffusion model. Thus, we could deal with scene images with arbitrary aspect ratios and boxes for extremely small or large areas.

\noindent\textbf{Benchmarks.}
For quantitative results, we construct a new benchmark with 30 new concepts provided by DreamBooth~\cite{dreambooth} for the target images. For the scene image, we manually pick 80 images with boxes in COCO-Val~\cite{coco}. Thus we generate 2,400 images for the object-scene combinations. 
We also make qualitative analysis on VitonHD-test~\cite{vitonhd} to validate the performance for virtual try-on.

\noindent\textbf{Evaluation metrics.}
On our constructed DreamBooth dataset, we follow DreamBooth~\cite{dreambooth} to calculate the CLIP-Score and DINO-Score, as these metrics could reflect the similarity between the generated region and the target object.  In addition,  we organize user studies with a group of 15 annotators to rate the generated results from the perspective of fidelity, quality, and diversity.

\subsection{Comparisons with Existing Alternatives}

\noindent\textbf{Reference-based methods.}
In \cref{fig:demobig}, we present the visualization results compared with previous reference-based methods. Paint-by-Example~\cite{paintbyexample} and Graphit~\cite{Graphit} support the same input format as ours, and they take a target image as input to edit a local region of a scene image without parameter tuning. IP-Adapter~\cite{ye2023ipadapter} is a universal method supporting image prompt, and we use its inpainting model for comparison.
We also compare Stable Diffusion~\cite{ldm}, which is a text-to-image model, and we use its inpainting version and give detailed text descriptions as the condition to conduct the generation for the text-described target.  

Results show that previous reference-based methods could only keep the semantic consistency with distinguishing features like the dog face on the backpack, and coarse granites of patterns like the color of the sloth toy. However, as those new concepts are not included in the training category, their generation results are far from ID-consistent. In contrast, our \method shows promising performance for zero-shot image customization with highly-faithful details.

\noindent\textbf{Tuning-based methods.}
Customized generation is extensively explored. Previous works~\cite{dreambooth,textinversion,liu2023cones,instantbooth,apprenticeshipsubject}  usually fine-tune a subject-specific text inversion to present the target object, thus making generations with arbitrary text prompts. They could better preserve the fidelity compared with previous reference-based methods, but have the following drawbacks: first, the fine-tuning usually requires 4-5 target images and takes nearly an hour; second, they could not specify the background scene and target locations; third, when it comes to multi-subject composition, the attributes of different subjects often mix together.  

In \cref{fig:multi}, we include tuning-based methods for comparisons and also use Paint-by-Example~\cite{paintbyexample} as the representative for previous reference-based methods.  Results show that Paint-by-Example~\cite{paintbyexample} performs well for trained categories like dog and cat~(in row~3) but performs poorly for new concepts~(row 1-2).  DreamBooth~
\cite{dreambooth}, Custom Diffusion~\cite{multiconcept}, and Cones~\cite{liu2023cones} give better fidelity for new concepts but still suffer from the problem of ``multi-subject confusion''. 
In contrast, \method owns the advantages of both reference- and tuning-based methods, which could generate high-fidelity results for mult-subject composition without the need for parameter tuning. 

\begin{table}[t]
\caption{%
    \textbf{User study} on the comparison between our \method and existing reference-based alternatives.
    ``Quality'', ``Fidelity'', and ``Diversity'' measure synthesis quality, object identity preservation, and object local variation (\textit{i.e.}, across four proposals), respectively.
    Each metric is rated from 1 (worst) to 4 (best).
}
\label{tab:userstudy}
\vspace{-7pt}
\centering\footnotesize
\setlength{\tabcolsep}{5.5pt}
\begin{tabular}{lccc}
\toprule
                                        & Quality~($\uparrow$) & Fidelity~($\uparrow$) & Diversity~($\uparrow$)  \\
\midrule
Paint-by-Example~\cite{paintbyexample}  & 2.71                 & 2.10                  & \textbf{3.04} \\
Graphit~\cite{Graphit}                  & 2.65                 & 2.11                  & 2.84 \\
\method (ours)                          & \textbf{3.04}        & \textbf{3.06}         & 2.88 \\
\bottomrule
\end{tabular}
\vspace{-10pt}
\end{table}

\noindent\textbf{User study.}
We organize a user study to compare Paint-by-Example~\cite{paintbyexample}, Graphit~\cite{Graphit}, and our model. We let 15 annotators rate 30 groups of images. For each group, we provide one target image and one scene image, and make each of the three models generates four predictions. We prepare detailed regulations and templates to rate the images for scores of 1 to 4 from three perspectives: ``Fidelity'', ``quality'', and ``diversity''. ``Fidelity'' measures the ability of ID preserving, and ``Quality'' counts for whether the generated image is harmonized without considering fidelity. As we do not encourage ``copy-paste'' style generation, we use ``diversity'' to measure the differences among the four generated proposals. The user-study results are listed in \cref{tab:userstudy}. It shows that our model owns obvious superiorities for fidelity and quantity, especially for fidelity.  However, as \cite{paintbyexample,Graphit} only keeps the semantic consistency, but our methods preserve the instance identity. They naturally have a larger space for diversity. In this case, \method still gets higher rates than \cite{Graphit} and competitive results with \cite{paintbyexample}, which verifies the effectiveness of our method.

\begin{figure}[t]
\centering 
\includegraphics[width=1.0\linewidth]{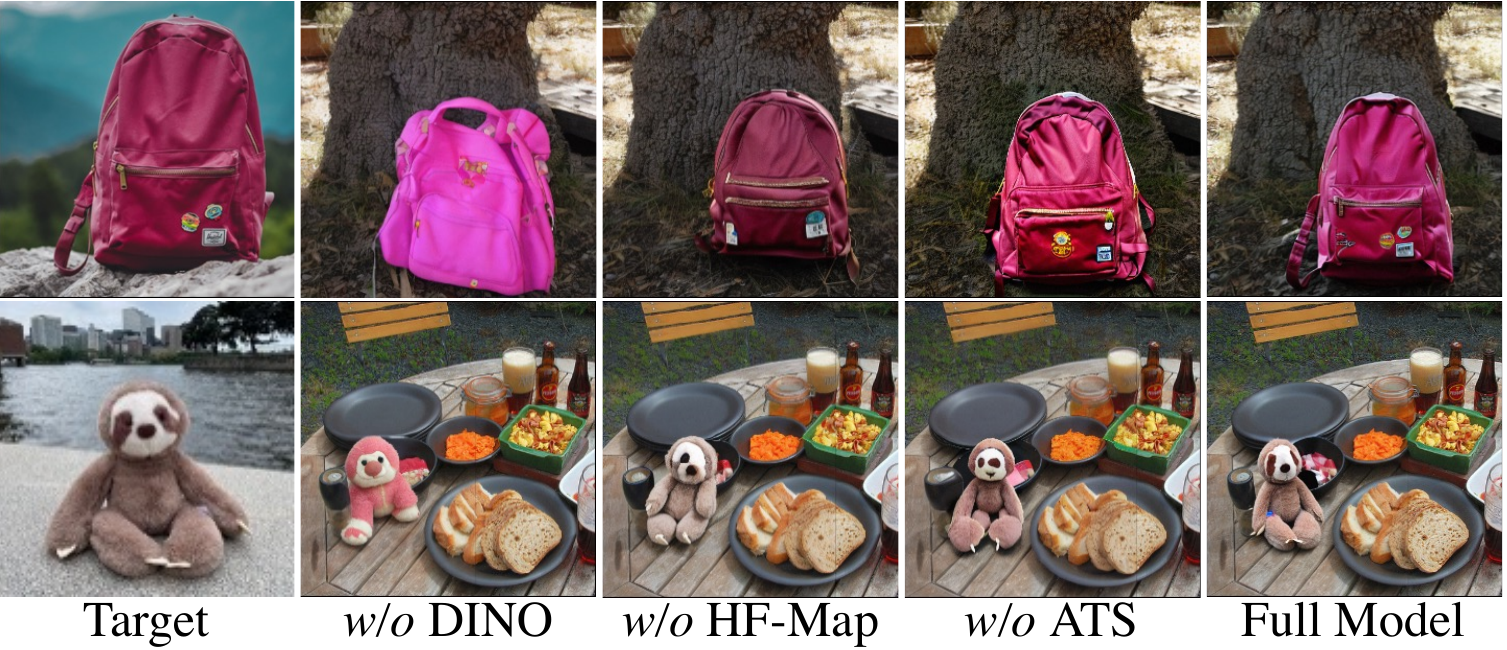} 
\vspace{-17pt}
\caption{%
    \textbf{Qualitative ablation studies} on the core components of \method.
    ``HF-Map'' stands for the high-frequency map in the detail extractor, while ``ATS'' refers to adaptive timestep sampling.
}
\label{fig:ablation_all}
\vspace{-5pt}
\end{figure}

\subsection{Ablation Studies}

We carry out extensive ablation studies to verify the effectiveness of our designs. We first validate the core components, then we dive into the details of the ID extractor and detail extractor to give an in-depth analysis.

\noindent\textbf{Core components.} 
As demonstrated in \cref{fig:ablation_all}, given the same target object, scene, and location, we analyze the generated results with different model designs. We demonstrate the generation results of \method in the last column and remove each core component individually to observe the influences.  We first change the backbone of our ID extractor from the DINOv2~\cite{dinov2} to CLIP image encoder~\cite{CLIP}, which is widely used in previous counterparts like \cite{paintbyexample,objectstitch}. We find the generated results lose the identity features, and could only keep the semantic consistency. 
Then, we set the collage region from the high-frequency map to an all-zero map like the inpainting baselines~\cite{controlnet,ldm}. We find that the fine details degenerate compared with our full model~(last column), like the logo of the bag~(row~1), and the eye shape of the toy sloth~(row~2). It shows that our frequency map effectively guides the generation of fine structural details.  
We also make ablation for our adaptive timestep sampling~(ATS) strategy. We replace ATS with an even distribution sampler and find the results present better diversity but are inferior for both image quality and fidelity.


\begin{figure}[t]
\centering 
\includegraphics[width=1.0\linewidth]{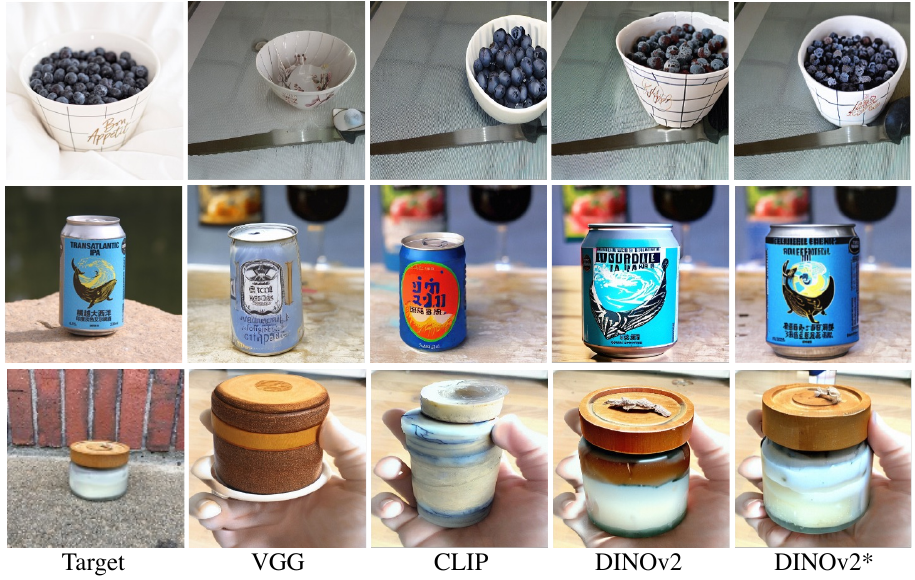} 
\vspace{-17pt}
\caption{%
    \textbf{Qualitative analysis of using different backbones for the ID extractor.}
    ``DINOv2*'' refers to removing the background of the target object with a frozen segmentation model before feeding it into the DINOv2 model.
}
\label{fig:idextractor}
\vspace{-5pt}
\end{figure}

\begin{table}[t]
\caption{%
    \textbf{Quantitative analysis of using different backbones for the ID extractor.}
    Here, ``G'' refers to the global token, ``P'' refers to patch tokens, and ``Seg'' refers to removing the background of the target object with a frozen segmentation model.
}
\label{tab:idextractor}
\vspace{-7pt}
\centering\footnotesize
\setlength{\tabcolsep}{9pt}
\begin{tabular}{lcc}
\toprule
    & CLIP Score~($\uparrow$) & DINO Score~($\uparrow$) \\
\midrule
VGG          & 71.7  & 27.7   \\
CLIP~(G+P)      & 73.8  &  31.5  \\
DINOv2~(G)        & 73.1 &  35.4  \\
DINOv2~(G+P)         & 81.0  &  64.1  \\
DINOv2~(G+P) + Seg         &  82.1 &  67.8  \\
\bottomrule
\end{tabular}
\vspace{-10pt}
\end{table}

\noindent\textbf{ID extractor.}
We explore the key factors for designing the ID extractor. In \cref{fig:idextractor}, we compare VGG~\cite{VGG}, CLIP~\cite{CLIP} and DINOv2~\cite{dinov2} to extract the ID tokens. We conclude that DINOv2~\cite{dinov2} shows a dominant superiority for keeping the target identity. We also verify that it is significant to filter out the background information for the target object, and DINOv2 could extract cleaner and more discriminative features. Quantitative results are listed in \cref{tab:idextractor}, which are consistent with our visual analysis.

\begin{figure}[t]
\centering 
\includegraphics[width=1.0\linewidth]{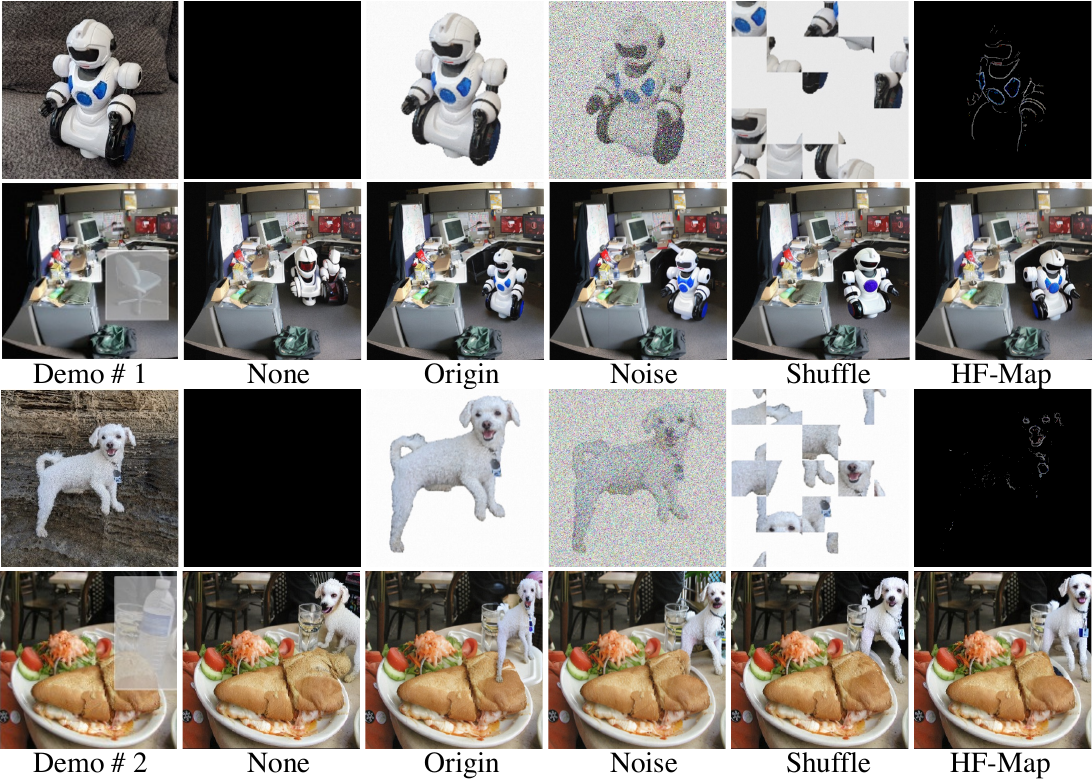} 
\vspace{-17pt}
\caption{%
    \textbf{Qualitative analysis of using different collages to extract details.}
    ``None'' means stitching the surroundings with an all-zero map.
    ``Origin'', ``Noise'', ``Shuffle'', and ``HF-Map'' refer to the original image with no background, noised image, patch-shuffled image, and the high-frequency map, respectively.
}
\label{fig:detailextractor}
\vspace{-5pt}
\end{figure}

\noindent\textbf{Detail extractor.}
We make multiple explorations for the collaged image. The CLIP and DINO scores are reported in \cref{tab:detailextractor}, compared with non-collage, all these collaging methods bring notable improvements. 
To make better comparisons, we give visualization results in \cref{fig:detailextractor}, which shows comparisons for no collage, pasting of the original target object, the noised inversion of the target object, the shuffled patches, and our high-frequency map. We observe a trade-off between fidelity and diversity. ``Original image'' presents the highest fidelity for both the robot and the dog, but the generated images seem like a copy-paste of the target. ``None'' shows the best diversity for the poses of the dog, but it lacks details like the badge of the dog and the whole shape of the robots. Among those methods, the high-frequency map shows a satisfactory trade-off, which keeps the majority of the details but adjusts the dog and robot with proper poses and views.

\begin{table}[t]
\caption{%
    \textbf{Quantitative analysis of using different collages to extract details.}
    It is noteworthy that, even the ``Original Image'' strategy best preserves the object identity, the object is with highly limited variation (\textit{i.e.}, almost with the same form as the target) in the synthesis. Hence, we pick HF-map as our standard setting.
}
\label{tab:detailextractor}
\vspace{-7pt}
\centering\footnotesize
\setlength{\tabcolsep}{9pt}
\begin{tabular}{lcc}
\toprule
Strategy  & CLIP Score~($\uparrow$) & DINO Score~($\uparrow$) \\
\midrule
None~(\textit{i.e.}, all-zero map)        & 80.4  & 63.2   \\
Original Image      & 82.2  & 68.8  \\
Noise Image  & 81.6  & 68.1  \\
Patch-shuffled Image        & 82.0  & 66.9  \\
High-frequency Map       & 82.1 &  67.8  \\
\bottomrule
\end{tabular}
\vspace{-10pt}
\end{table}

\subsection{More Applications}

\begin{figure}[t]
\centering 
\includegraphics[width=1.0\linewidth]{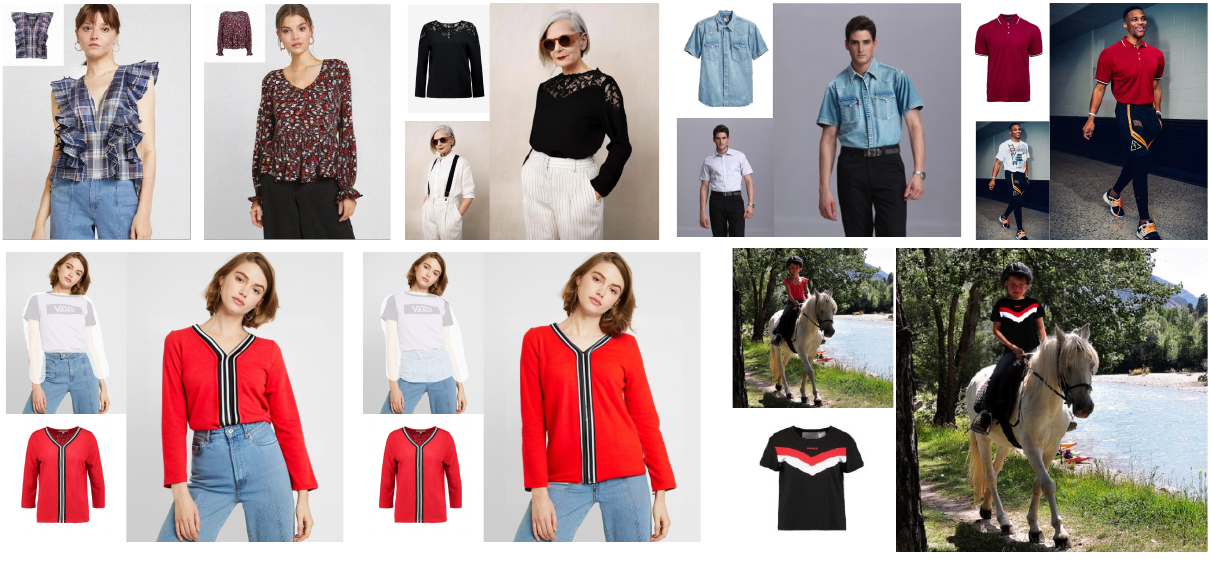} 
\vspace{-20pt}
\caption{%
    \textbf{Performance of \method on virtual try-on} on VitonHD-test~\cite{vitonhd} and real-life scenarios. \method could preserve the color, texture, and patterns of the target clothes and customize the garment shape~(bottom left) with mask control. 
}
\label{fig:tryon}
\vspace{-4pt}
\end{figure}

\noindent\textbf{Virtual try-on.}
As shown in \cref{fig:tryon}, without task-specific tuning, \method could give satisfactory performance for virtual try-on on VitonHD-test~\cite{vitonhd} and real-world scenarios for human with different sexes, ages, and races. Besides, AnyDoor supports users to draw coarse contour maps to control the style like tuck in or untuck~(second row left) and shows strong generalization abilities for real-life scenarios with complex backgrounds.

\begin{figure}[t]
\centering
\includegraphics[width=1.0\linewidth]{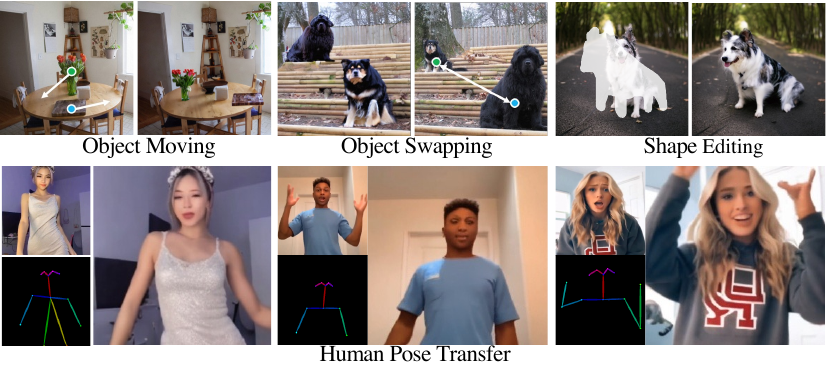} 
\vspace{-18pt}
\caption{%
    \textbf{Demonstrations for more applications of \method}, such as object moving, object swapping, shape editing, and human pose transfer. In row~2, the reference images and the novel poses are presented on the left with the generated results on the right.
}
\label{fig:drag}
\vspace{-12pt}
\end{figure}

\noindent\textbf{Extensible controls.}
As demonstrated in \cref{fig:drag}, it is easy to extend \method to realize more fantastic functions like object moving, swapping, and reshaping. When taking a pose skeleton map as an additional control, \method could even serve as a strong baseline for human pose transfer.

The pipeline of object moving, swapping, and reshaping incorporates an additional inpainting model~\cite{ldm} and an interactive segmentation model~\cite{segmentanything}. We first get the mask of the object by clicking. Then, we use the inpainting model to fill the object's original position according to the scene background and apply the \method to re-generate it at the new location with optional shape control.

In the second row of \cref{fig:drag}, when conducting human pose transfer, we add an extra ControlNet-copy on \method to control the human pose.  Then, we train the model with the same configuration of Disco~\cite{wang2023disco}, a state-of-the-art human pose transfer method.  The results are impressive that \method keeps the identity well for both the human faces and garments.

\section{Conclusion}
\vspace{-5pt}
We present \method for object teleportation. The core idea is to use a discriminative ID extractor and a frequency-aware detail extractor to characterize the target object. Trained on a large combination of video and image data, we composite the object at the specific location of the scene image with effective shape control. \method provides a universal solution for general region-to-region mapping tasks and could be profitable for various applications. 

\noindent \textbf{Limitations.}
\method shows impressive results for keeping the object identification. However, it still struggles with fine details like the small characters or logos. This issue might be solved by collecting related training data, enlarging the resolution, and training better VAE decoders.

{
\small
\bibliographystyle{ieeenat_fullname}
\bibliography{ref.bib}

\begin{thebibliography}{64}
\providecommand{\natexlab}[1]{#1}
\providecommand{\url}[1]{\texttt{#1}}
\expandafter\ifx\csname urlstyle\endcsname\relax
  \providecommand{\doi}[1]{doi: #1}\else
  \providecommand{\doi}{doi: \begingroup \urlstyle{rm}\Url}\fi

\bibitem[Athar et~al.(2023)Athar, Luiten, Voigtlaender, Khurana, Dave, Leibe, and Ramanan]{burst}
Ali Athar, Jonathon Luiten, Paul Voigtlaender, Tarasha Khurana, Achal Dave, Bastian Leibe, and Deva Ramanan.
\newblock Burst: A benchmark for unifying object recognition, segmentation and tracking in video.
\newblock In \emph{WACV}, 2023.

\bibitem[Avrahami et~al.(2022)Avrahami, Lischinski, and Fried]{blendeddiffusion}
Omri Avrahami, Dani Lischinski, and Ohad Fried.
\newblock Blended diffusion for text-driven editing of natural images.
\newblock In \emph{CVPR}, 2022.

\bibitem[Avrahami et~al.(2023)Avrahami, Aberman, Fried, Cohen-Or, and Lischinski]{Break-A-Scene}
Omri Avrahami, Kfir Aberman, Ohad Fried, Daniel Cohen-Or, and Dani Lischinski.
\newblock Break-a-scene: Extracting multiple concepts from a single image.
\newblock In \emph{SIGGRAPH Asia}, 2023.

\bibitem[Borji et~al.(2015)Borji, Cheng, Jiang, and Li]{MSRA10K}
Ali Borji, Ming-Ming Cheng, Huaizu Jiang, and Jia Li.
\newblock Salient object detection: A benchmark.
\newblock \emph{TIP}, 2015.

\bibitem[Cao et~al.(2023)Cao, Wang, Qi, Shan, Qie, and Zheng]{masactrl}
Mingdeng Cao, Xintao Wang, Zhongang Qi, Ying Shan, Xiaohu Qie, and Yinqiang Zheng.
\newblock Masactrl: Tuning-free mutual self-attention control for consistent image synthesis and editing.
\newblock In \emph{ICCV}, 2023.

\bibitem[Casanova et~al.(2023)Casanova, Careil, Romero-Soriano, Pal, Verbeek, and Drozdzal]{controllablecollage}
Arantxa Casanova, Marl{\`e}ne Careil, Adriana Romero-Soriano, Christopher~J Pal, Jakob Verbeek, and Michal Drozdzal.
\newblock Controllable image generation via collage representations.
\newblock \emph{arXiv:2304.13722}, 2023.

\bibitem[Chen and Kae(2019)]{towardharm}
Bor-Chun Chen and Andrew Kae.
\newblock Toward realistic image compositing with adversarial learning.
\newblock In \emph{CVPR}, 2019.

\bibitem[Chen et~al.(2022{\natexlab{a}})Chen, Gu, Li, Lan, Meng, Wang, and Li]{hierarchicalharm}
Haoxing Chen, Zhangxuan Gu, Yaohui Li, Jun Lan, Changhua Meng, Weiqiang Wang, and Huaxiong Li.
\newblock Hierarchical dynamic image harmonization.
\newblock In \emph{ACMMM}, 2022{\natexlab{a}}.

\bibitem[Chen et~al.(2023{\natexlab{a}})Chen, Zhang, Wang, Duan, Zhou, and Zhu]{disenbooth}
Hong Chen, Yipeng Zhang, Xin Wang, Xuguang Duan, Yuwei Zhou, and Wenwu Zhu.
\newblock Disenbooth: Disentangled parameter-efficient tuning for subject-driven text-to-image generation.
\newblock \emph{arXiv:2305.03374}, 2023{\natexlab{a}}.

\bibitem[Chen et~al.(2023{\natexlab{b}})Chen, Hu, Li, Rui, Jia, Chang, and Cohen]{apprenticeshipsubject}
Wenhu Chen, Hexiang Hu, Yandong Li, Nataniel Rui, Xuhui Jia, Ming-Wei Chang, and William~W Cohen.
\newblock Subject-driven text-to-image generation via apprenticeship learning.
\newblock In \emph{NeurIPS}, 2023{\natexlab{b}}.

\bibitem[Chen et~al.(2021)Chen, Zhao, Yu, Zhang, and Duan]{cdnet}
Xi Chen, Zhiyan Zhao, Feiwu Yu, Yilei Zhang, and Manni Duan.
\newblock Conditional diffusion for interactive segmentation.
\newblock In \emph{ICCV}, 2021.

\bibitem[Chen et~al.(2022{\natexlab{b}})Chen, Zhao, Zhang, Duan, Qi, and Zhao]{focalclick}
Xi Chen, Zhiyan Zhao, Yilei Zhang, Manni Duan, Donglian Qi, and Hengshuang Zhao.
\newblock Focalclick: towards practical interactive image segmentation.
\newblock In \emph{CVPR}, 2022{\natexlab{b}}.

\bibitem[Choi et~al.(2021)Choi, Park, Lee, and Choo]{vitonhd}
Seunghwan Choi, Sunghyun Park, Minsoo Lee, and Jaegul Choo.
\newblock Viton-hd: High-resolution virtual try-on via misalignment-aware normalization.
\newblock In \emph{CVPR}, 2021.

\bibitem[Cong et~al.(2020)Cong, Zhang, Niu, Liu, Ling, Li, and Zhang]{dovenetharm}
Wenyan Cong, Jianfu Zhang, Li Niu, Liu Liu, Zhixin Ling, Weiyuan Li, and Liqing Zhang.
\newblock Dovenet: Deep image harmonization via domain verification.
\newblock In \emph{CVPR}, 2020.

\bibitem[Cong et~al.(2022)Cong, Tao, Niu, Liang, Gao, Sun, and Zhang]{dualtrasnharm}
Wenyan Cong, Xinhao Tao, Li Niu, Jing Liang, Xuesong Gao, Qihao Sun, and Liqing Zhang.
\newblock High-resolution image harmonization via collaborative dual transformations.
\newblock In \emph{CVPR}, 2022.

\bibitem[Contributors(2023)]{Graphit}
Graphit Contributors.
\newblock Graphit: A unified framework for diverse image editing tasks.
\newblock \url{https://github.com/navervision/Graphit}, 2023.

\bibitem[Ding et~al.(2023)Ding, Liu, He, Jiang, Torr, and Bai]{ding2023mose}
Henghui Ding, Chang Liu, Shuting He, Xudong Jiang, Philip~HS Torr, and Song Bai.
\newblock Mose: A new dataset for video object segmentation in complex scenes.
\newblock In \emph{ICCV}, 2023.

\bibitem[Gal et~al.(2023)Gal, Alaluf, Atzmon, Patashnik, Bermano, Chechik, and Cohen-Or]{textinversion}
Rinon Gal, Yuval Alaluf, Yuval Atzmon, Or Patashnik, Amit~H Bermano, Gal Chechik, and Daniel Cohen-Or.
\newblock An image is worth one word: Personalizing text-to-image generation using textual inversion.
\newblock In \emph{ICLR}, 2023.

\bibitem[Gu et~al.(2023)Gu, Wang, Wu, Shi, Chen, Fan, Xiao, Zhao, Chang, Wu, et~al.]{Mix-of-Show}
Yuchao Gu, Xintao Wang, Jay~Zhangjie Wu, Yujun Shi, Yunpeng Chen, Zihan Fan, Wuyou Xiao, Rui Zhao, Shuning Chang, Weijia Wu, et~al.
\newblock Mix-of-show: Decentralized low-rank adaptation for multi-concept customization of diffusion models.
\newblock In \emph{NeurIPS}, 2023.

\bibitem[Guo et~al.(2021)Guo, Zheng, Jiang, Gu, and Zheng]{intrinsicharm}
Zonghui Guo, Haiyong Zheng, Yufeng Jiang, Zhaorui Gu, and Bing Zheng.
\newblock Intrinsic image harmonization.
\newblock In \emph{CVPR}, 2021.

\bibitem[Gupta et~al.(2019)Gupta, Dollar, and Girshick]{lvis}
Agrim Gupta, Piotr Dollar, and Ross Girshick.
\newblock Lvis: A dataset for large vocabulary instance segmentation.
\newblock In \emph{CVPR}, 2019.

\bibitem[Huang et~al.(2023)Huang, Chen, Liu, Shen, Zhao, and Zhou]{composer}
Lianghua Huang, Di Chen, Yu Liu, Yujun Shen, Deli Zhao, and Jingren Zhou.
\newblock Composer: Creative and controllable image synthesis with composable conditions.
\newblock In \emph{ICML}, 2023.

\bibitem[Kanopoulos et~al.(1988)Kanopoulos, Vasanthavada, and Baker]{sobel}
Nick Kanopoulos, Nagesh Vasanthavada, and Robert~L Baker.
\newblock Design of an image edge detection filter using the sobel operator.
\newblock \emph{JSSC}, 1988.

\bibitem[Kawar et~al.(2023)Kawar, Zada, Lang, Tov, Chang, Dekel, Mosseri, and Irani]{imagic}
Bahjat Kawar, Shiran Zada, Oran Lang, Omer Tov, Huiwen Chang, Tali Dekel, Inbar Mosseri, and Michal Irani.
\newblock Imagic: Text-based real image editing with diffusion models.
\newblock In \emph{CVPR}, 2023.

\bibitem[Kingma and Ba(2014)]{adam}
Diederik~P Kingma and Jimmy Ba.
\newblock Adam: A method for stochastic optimization.
\newblock \emph{arXiv:1412.6980}, 2014.

\bibitem[Kirillov et~al.(2023)Kirillov, Mintun, Ravi, Mao, Rolland, Gustafson, Xiao, Whitehead, Berg, Lo, et~al.]{segmentanything}
Alexander Kirillov, Eric Mintun, Nikhila Ravi, Hanzi Mao, Chloe Rolland, Laura Gustafson, Tete Xiao, Spencer Whitehead, Alexander~C Berg, Wan-Yen Lo, et~al.
\newblock Segment anything.
\newblock In \emph{ICCV}, 2023.

\bibitem[Kumari et~al.(2023)Kumari, Zhang, Zhang, Shechtman, and Zhu]{multiconcept}
Nupur Kumari, Bingliang Zhang, Richard Zhang, Eli Shechtman, and Jun-Yan Zhu.
\newblock Multi-concept customization of text-to-image diffusion.
\newblock In \emph{CVPR}, 2023.

\bibitem[Li et~al.(2023{\natexlab{a}})Li, Li, and Hoi]{blipdiffusion}
Dongxu Li, Junnan Li, and Steven~CH Hoi.
\newblock Blip-diffusion: Pre-trained subject representation for controllable text-to-image generation and editing.
\newblock In \emph{NeurIPS}, 2023{\natexlab{a}}.

\bibitem[Li et~al.(2023{\natexlab{b}})Li, Li, Savarese, and Hoi]{blip2}
Junnan Li, Dongxu Li, Silvio Savarese, and Steven Hoi.
\newblock Blip-2: Bootstrapping language-image pre-training with frozen image encoders and large language models.
\newblock In \emph{ICML}, 2023{\natexlab{b}}.

\bibitem[Li et~al.(2023{\natexlab{c}})Li, Ku, Wei, and Chen]{li2023dreamedit}
Tianle Li, Max Ku, Cong Wei, and Wenhu Chen.
\newblock Dreamedit: Subject-driven image editing.
\newblock \emph{arXiv:2306.12624}, 2023{\natexlab{c}}.

\bibitem[Lin et~al.(2014)Lin, Maire, Belongie, Hays, Perona, Ramanan, Doll{\'a}r, and Zitnick]{coco}
Tsung-Yi Lin, Michael Maire, Serge Belongie, James Hays, Pietro Perona, Deva Ramanan, Piotr Doll{\'a}r, and C~Lawrence Zitnick.
\newblock Microsoft coco: Common objects in context.
\newblock In \emph{ECCV}, 2014.

\bibitem[Liu et~al.(2023{\natexlab{a}})Liu, Xu, Bertasius, and Niethammer]{simpleclick}
Qin Liu, Zhenlin Xu, Gedas Bertasius, and Marc Niethammer.
\newblock Simpleclick: Interactive image segmentation with simple vision transformers.
\newblock In \emph{ICCV}, 2023{\natexlab{a}}.

\bibitem[Liu et~al.(2023{\natexlab{b}})Liu, Feng, Zhu, Zhang, Zheng, Liu, Zhao, Zhou, and Cao]{liu2023cones}
Zhiheng Liu, Ruili Feng, Kai Zhu, Yifei Zhang, Kecheng Zheng, Yu Liu, Deli Zhao, Jingren Zhou, and Yang Cao.
\newblock Cones: Concept neurons in diffusion models for customized generation.
\newblock In \emph{ICML}, 2023{\natexlab{b}}.

\bibitem[Liu et~al.(2023{\natexlab{c}})Liu, Zhang, Shen, Zheng, Zhu, Feng, Liu, Zhao, Zhou, and Cao]{cones2}
Zhiheng Liu, Yifei Zhang, Yujun Shen, Kecheng Zheng, Kai Zhu, Ruili Feng, Yu Liu, Deli Zhao, Jingren Zhou, and Yang Cao.
\newblock Cones 2: Customizable image synthesis with multiple subjects.
\newblock In \emph{NeurIPS}, 2023{\natexlab{c}}.

\bibitem[Miao et~al.(2022)Miao, Wang, Wu, Li, Zhang, Wei, and Yang]{vipseg}
Jiaxu Miao, Xiaohan Wang, Yu Wu, Wei Li, Xu Zhang, Yunchao Wei, and Yi Yang.
\newblock Large-scale video panoptic segmentation in the wild: A benchmark.
\newblock In \emph{CVPR}, 2022.

\bibitem[Oquab et~al.(2024)Oquab, Darcet, Moutakanni, Vo, Szafraniec, Khalidov, Fernandez, Haziza, Massa, El-Nouby, et~al.]{dinov2}
Maxime Oquab, Timoth{\'e}e Darcet, Th{\'e}o Moutakanni, Huy Vo, Marc Szafraniec, Vasil Khalidov, Pierre Fernandez, Daniel Haziza, Francisco Massa, Alaaeldin El-Nouby, et~al.
\newblock Dinov2: Learning robust visual features without supervision.
\newblock \emph{TMLR}, 2024.

\bibitem[Qin et~al.(2023)Qin, Zhang, Yu, Feng, Yang, Zhou, Wang, Niebles, Xiong, Savarese, et~al.]{unicontrol}
Can Qin, Shu Zhang, Ning Yu, Yihao Feng, Xinyi Yang, Yingbo Zhou, Huan Wang, Juan~Carlos Niebles, Caiming Xiong, Silvio Savarese, et~al.
\newblock Unicontrol: A unified diffusion model for controllable visual generation in the wild.
\newblock In \emph{NeurIPS}, 2023.

\bibitem[Qin et~al.(2020)Qin, Zhang, Huang, Dehghan, Zaiane, and Jagersand]{u2net}
Xuebin Qin, Zichen Zhang, Chenyang Huang, Masood Dehghan, Osmar~R Zaiane, and Martin Jagersand.
\newblock U2-net: Going deeper with nested u-structure for salient object detection.
\newblock \emph{PR}, 2020.

\bibitem[Radford et~al.(2021)Radford, Kim, Hallacy, Ramesh, Goh, Agarwal, Sastry, Askell, Mishkin, Clark, et~al.]{CLIP}
Alec Radford, Jong~Wook Kim, Chris Hallacy, Aditya Ramesh, Gabriel Goh, Sandhini Agarwal, Girish Sastry, Amanda Askell, Pamela Mishkin, Jack Clark, et~al.
\newblock Learning transferable visual models from natural language supervision.
\newblock In \emph{ICML}, 2021.

\bibitem[Ramesh et~al.(2022)Ramesh, Dhariwal, Nichol, Chu, and Chen]{DALLE2}
Aditya Ramesh, Prafulla Dhariwal, Alex Nichol, Casey Chu, and Mark Chen.
\newblock Hierarchical text-conditional image generation with clip latents.
\newblock \emph{arXiv:2204.06125}, 2022.

\bibitem[Rombach et~al.(2022)Rombach, Blattmann, Lorenz, Esser, and Ommer]{ldm}
Robin Rombach, Andreas Blattmann, Dominik Lorenz, Patrick Esser, and Bj{\"o}rn Ommer.
\newblock High-resolution image synthesis with latent diffusion models.
\newblock In \emph{CVPR}, 2022.

\bibitem[Ruiz et~al.(2023)Ruiz, Li, Jampani, Pritch, Rubinstein, and Aberman]{dreambooth}
Nataniel Ruiz, Yuanzhen Li, Varun Jampani, Yael Pritch, Michael Rubinstein, and Kfir Aberman.
\newblock Dreambooth: Fine tuning text-to-image diffusion models for subject-driven generation.
\newblock In \emph{CVPR}, 2023.

\bibitem[Saharia et~al.(2022)Saharia, Chan, Saxena, Li, Whang, Denton, Ghasemipour, Gontijo~Lopes, Karagol~Ayan, Salimans, et~al.]{imagen}
Chitwan Saharia, William Chan, Saurabh Saxena, Lala Li, Jay Whang, Emily~L Denton, Kamyar Ghasemipour, Raphael Gontijo~Lopes, Burcu Karagol~Ayan, Tim Salimans, et~al.
\newblock Photorealistic text-to-image diffusion models with deep language understanding.
\newblock In \emph{NeurIPS}, 2022.

\bibitem[Sarukkai et~al.(2024)Sarukkai, Li, Ma, R{\'e}, and Fatahalian]{Collagediffusion}
Vishnu Sarukkai, Linden Li, Arden Ma, Christopher R{\'e}, and Kayvon Fatahalian.
\newblock Collage diffusion.
\newblock In \emph{WACV}, 2024.

\bibitem[Shi et~al.(2023)Shi, Xiong, Lin, and Jung]{instantbooth}
Jing Shi, Wei Xiong, Zhe Lin, and Hyun~Joon Jung.
\newblock Instantbooth: Personalized text-to-image generation without test-time finetuning.
\newblock \emph{arXiv:2304.03411}, 2023.

\bibitem[Simonyan and Zisserman(2014)]{VGG}
Karen Simonyan and Andrew Zisserman.
\newblock Very deep convolutional networks for large-scale image recognition.
\newblock \emph{arXiv:1409.1556}, 2014.

\bibitem[Song et~al.(2023)Song, Zhang, Lin, Cohen, Price, Zhang, Kim, and Aliaga]{objectstitch}
Yizhi Song, Zhifei Zhang, Zhe Lin, Scott Cohen, Brian Price, Jianming Zhang, Soo~Ye Kim, and Daniel Aliaga.
\newblock Objectstitch: Object compositing with diffusion model.
\newblock In \emph{CVPR}, 2023.

\bibitem[Sunkavalli et~al.(2010)Sunkavalli, Johnson, Matusik, and Pfister]{multiharm}
Kalyan Sunkavalli, Micah~K Johnson, Wojciech Matusik, and Hanspeter Pfister.
\newblock Multi-scale image harmonization.
\newblock In \emph{SIGGRAPH}, 2010.

\bibitem[Wang et~al.(2017)Wang, Lu, Wang, Feng, Wang, Yin, and Ruan]{DUT}
Lijun Wang, Huchuan Lu, Yifan Wang, Mengyang Feng, Dong Wang, Baocai Yin, and Xiang Ruan.
\newblock Learning to detect salient objects with image-level supervision.
\newblock In \emph{CVPR}, 2017.

\bibitem[Wang et~al.(2023)Wang, Li, Lin, Lin, Yang, Zhang, Liu, and Wang]{wang2023disco}
Tan Wang, Linjie Li, Kevin Lin, Chung-Ching Lin, Zhengyuan Yang, Hanwang Zhang, Zicheng Liu, and Lijuan Wang.
\newblock Disco: Disentangled control for referring human dance generation in real world.
\newblock \emph{arXiv:2307.00040}, 2023.

\bibitem[Wang et~al.(2021)Wang, Feiszli, Wang, and Tran]{UVO}
Weiyao Wang, Matt Feiszli, Heng Wang, and Du Tran.
\newblock Unidentified video objects: A benchmark for dense, open-world segmentation.
\newblock In \emph{ICCV}, 2021.

\bibitem[Xiao et~al.(2023)Xiao, Yin, Freeman, Durand, and Han]{fastcomposer}
Guangxuan Xiao, Tianwei Yin, William~T Freeman, Fr{\'e}do Durand, and Song Han.
\newblock Fastcomposer: Tuning-free multi-subject image generation with localized attention.
\newblock \emph{arXiv:2305.10431}, 2023.

\bibitem[Xie et~al.(2023)Xie, Zhang, Lin, Hinz, and Zhang]{smartbrush}
Shaoan Xie, Zhifei Zhang, Zhe Lin, Tobias Hinz, and Kun Zhang.
\newblock Smartbrush: Text and shape guided object inpainting with diffusion model.
\newblock In \emph{CVPR}, 2023.

\bibitem[Xu et~al.(2018)Xu, Yang, Fan, Yue, Liang, Yang, and Huang]{youtubevos}
Ning Xu, Linjie Yang, Yuchen Fan, Dingcheng Yue, Yuchen Liang, Jianchao Yang, and Thomas Huang.
\newblock Youtube-vos: A large-scale video object segmentation benchmark.
\newblock \emph{arXiv:1809.03327}, 2018.

\bibitem[Xue et~al.(2022)Xue, Ran, Chen, Jia, Zhao, and Tang]{dccf}
Ben Xue, Shenghui Ran, Quan Chen, Rongfei Jia, Binqiang Zhao, and Xing Tang.
\newblock Dccf: Deep comprehensible color filter learning framework for high-resolution image harmonization.
\newblock In \emph{ECCV}, 2022.

\bibitem[Yang et~al.(2023)Yang, Gu, Zhang, Zhang, Chen, Sun, Chen, and Wen]{paintbyexample}
Binxin Yang, Shuyang Gu, Bo Zhang, Ting Zhang, Xuejin Chen, Xiaoyan Sun, Dong Chen, and Fang Wen.
\newblock Paint by example: Exemplar-based image editing with diffusion models.
\newblock In \emph{CVPR}, 2023.

\bibitem[Yang et~al.(2019)Yang, Fan, and Xu]{youtubevis}
Linjie Yang, Yuchen Fan, and Ning Xu.
\newblock Video instance segmentation.
\newblock In \emph{ICCV}, 2019.

\bibitem[Ye et~al.(2023)Ye, Zhang, Liu, Han, and Yang]{ye2023ipadapter}
Hu Ye, Jun Zhang, Sibo Liu, Xiao Han, and Wei Yang.
\newblock Ip-adapter: Text compatible image prompt adapter for text-to-image diffusion models.
\newblock \emph{arXiv:2308.06721}, 2023.

\bibitem[Yu et~al.(2023{\natexlab{a}})Yu, Feng, Feng, Liu, Jin, Zeng, and Chen]{inpaintanything}
Tao Yu, Runseng Feng, Ruoyu Feng, Jinming Liu, Xin Jin, Wenjun Zeng, and Zhibo Chen.
\newblock Inpaint anything: Segment anything meets image inpainting.
\newblock \emph{arXiv:2304.06790}, 2023{\natexlab{a}}.

\bibitem[Yu et~al.(2023{\natexlab{b}})Yu, Xu, Zhang, Liu, Ye, Wu, Yan, Zhu, Xiong, Liang, et~al.]{mvimgnet}
Xianggang Yu, Mutian Xu, Yidan Zhang, Haolin Liu, Chongjie Ye, Yushuang Wu, Zizheng Yan, Chenming Zhu, Zhangyang Xiong, Tianyou Liang, et~al.
\newblock Mvimgnet: A large-scale dataset of multi-view images.
\newblock In \emph{CVPR}, 2023{\natexlab{b}}.

\bibitem[Yuan et~al.(2023)Yuan, Cao, Wang, Qi, Yuan, and Shan]{yuan2023customnet}
Ziyang Yuan, Mingdeng Cao, Xintao Wang, Zhongang Qi, Chun Yuan, and Ying Shan.
\newblock Customnet: Zero-shot object customization with variable-viewpoints in text-to-image diffusion models.
\newblock \emph{arXiv:2310.19784}, 2023.

\bibitem[Zhang and Agrawala(2023)]{controlnet}
Lvmin Zhang and Maneesh Agrawala.
\newblock Adding conditional control to text-to-image diffusion models.
\newblock In \emph{ICCV}, 2023.

\bibitem[Zhang et~al.(2023)Zhang, Han, Ghosh, Metaxas, and Ren]{sine}
Zhixing Zhang, Ligong Han, Arnab Ghosh, Dimitris~N Metaxas, and Jian Ren.
\newblock Sine: Single image editing with text-to-image diffusion models.
\newblock In \emph{CVPR}, 2023.

\bibitem[Zheng et~al.(2019)Zheng, Song, Chen, Hu, Cao, and Nie]{fahiontryon}
Na Zheng, Xuemeng Song, Zhaozheng Chen, Linmei Hu, Da Cao, and Liqiang Nie.
\newblock Virtually trying on new clothing with arbitrary poses.
\newblock In \emph{ACMMM}, 2019.

\end{thebibliography}
}

\end{document}